\documentclass[10pt,twocolumn,letterpaper]{article}

\usepackage{graphicx}
\usepackage{amsmath}
\usepackage{amssymb}
\usepackage{booktabs}
\usepackage{multirow}
\usepackage{diagbox}
\usepackage{color}

\usepackage[pagebackref,breaklinks,colorlinks]{hyperref}

\usepackage[capitalize]{cleveref}
\crefname{section}{Sec.}{Secs.}
\Crefname{section}{Section}{Sections}
\Crefname{table}{Table}{Tables}
\crefname{table}{Tab.}{Tabs.}

\begin{document}

\title{Zero-shot Image Captioning by Anchor-augmented Vision-Language Space Alignment}

\author{Junyang Wang
\and
Yi Zhang
\and
Ming Yan
\and
Ji Zhang
\and
Jitao Sang
}
\maketitle

\begin{abstract}
CLIP (Contrastive Language-Image Pre-Training) has shown remarkable zero-shot transfer capabilities in cross-modal correlation tasks such as visual classification and image retrieval. However, its performance in cross-modal generation tasks like zero-shot image captioning remains unsatisfied. In this work, we discuss that directly employing CLIP for zero-shot image captioning relies more on the textual modality in context and largely ignores the visual information, which we call \emph{contextual language prior}. To address this, we propose Cross-modal Language Models (CLMs) to facilitate unsupervised cross-modal learning. We further propose Anchor Augment to guide the generative model's attention to the fine-grained information in the representation of CLIP. Experiments on MS COCO and Flickr 30K validate the promising performance of proposed approach in both captioning quality and computational efficiency.
\end{abstract}

\section{Introduction}


Vision-Language Pre-training (VLP) has advanced the research of multi-modal modeling in recent years \cite{tan2019lxmert, chen2019uniter, li2020oscar, li2021align}, among which CLIP \cite{radford2021learning} has drawn increasing attention for its transferable visual representation learning. Benefiting from contrastive learning on a large-scale web image-text dataset, CLIP independently encodes images and text and maps them into a vision-language space with common semantics, thus making the zero-shot transfer between the two modalities possible \cite{xu2021simple, zhong2022regionCLIP, shi2022proposalCLIP, ramesh2022hierarchical, yu2022towards, esmaeilpour2022zero}. Impressive zero-shot image classification capability (76.2$\%$ accuracy on ImageNet) was demonstrated by CLIP \cite{radford2021learning}. 

\begin{figure}
\centering
\includegraphics[width=0.45 \textwidth]{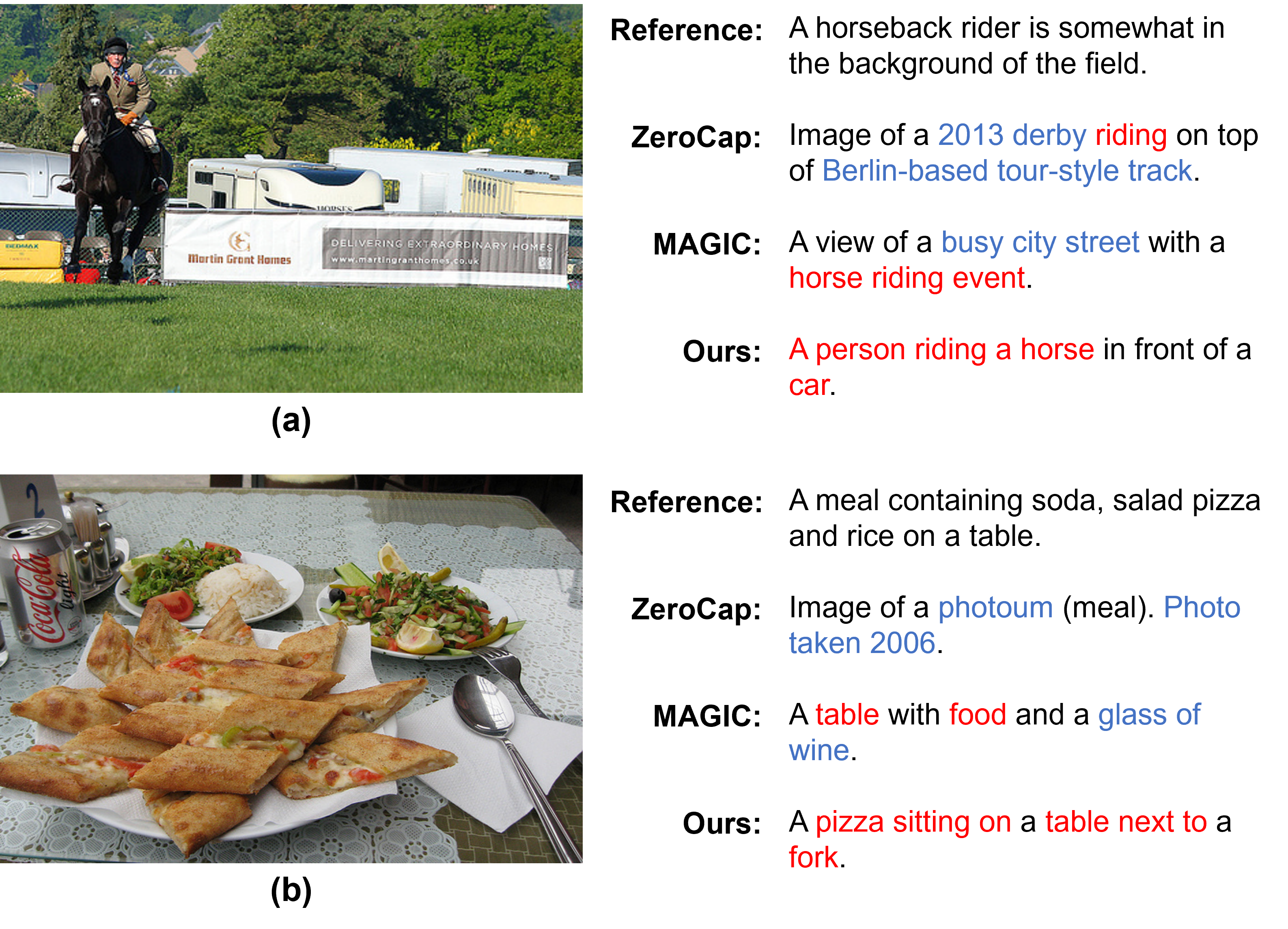}
\caption{\label{fig:sample} We show different captions generated by three zero-shot image captioning approaches: ZeroCap \cite{tewel2022zerocap}, MAGIC \cite{su2022language}, and ours. The red-colored words refer to relevant information, while the blue-colored words refer to irrelevant information that cannot be inferred from the image.
Observations include: (1) ZeroCap fails to capture the relevant information due to the overpowering contextual language prior; (2) MAGIC manages to encode the relevant information, but generates irrelevant information such as the ``street'' due to ``busy'' and the ``glass of wine'' due to ``table'' and ``food'' via ungrounded correlation from contextual language.}
\end{figure}

The zero-shot classification ability of CLIP has encouraged research on zero-shot image captioning. Existing CLIP-based zero-shot image captioning approaches \cite{tewel2022zerocap,su2022language} use a language model by the means of next-token prediction method to first suggest candidate words and then calculate the representation similarities of CLIP between each candidate word and image to select the generated word. These approaches use two constraints: one on constraining the generated caption to be as similar as possible to the given image in the CLIP representation space, and the other on constraining the generated caption to have as low a loss as possible on the language model. It is easy to see that the second constraint is only imposed on the textual modality. We discuss that this uni-modal candidate word-centric solution is prone to produce \emph{contextual language prior}: The language model suggests candidate words only based on the context of the generated caption, by exploiting the prior activated in the language model. While image captioning is essentially a cross-modal task, the above contextual language prior issue can make the caption generation ignore the visual information of the given image and thus generates irrelevant content as illustrated in Figure \ref{fig:sample}.

Cross-modal learning is thus needed in CLIP-based zero-shot image captioning. The problem turns to address the unavailability of supervised cross-modal data under zero-shot settings. We are inspired by the fact that CLIP aligns visual and textual representations in cross-modal embedding space, and propose Cross-modal Language Models (CLMs) to transform uni-modal data in cross-modal representations to facilitate unsupervised cross-modal learning. Specifically, we first use CLIP to obtain the representations of the sentences in the unsupervised corpus and place the representations in the first position of the language model as prefix tokens. And then we use the original sentences as self-supervised labels for auto-regressive language modeling training. 

With the cross-modal learning of CLIP representation, CLMs manages to employ both textual and visual modalities for caption generation at a global level. However, the fine-grained information in cross-modal representation of CLIP is still not fully exploited. To improve the attention of the generative model to the fine-grained information, we propose Anchor Augment. In the training phase, we extract nouns from the sentences as anchors and use them for CLMs. We hope that anchors can be used as cues for the fine-grained information in the representation of CLIP. To improve the robustness of anchors, we further introduce anchor random dropout: dropping out all anchors in a training sample with a certain probability. In the generation phase, we use the object detector to extract the labels of the objects in the images as anchors, which improves the quality of caption generation.


We summarize the contributions as follows:
\begin{itemize}
\item We position the contextual language prior issue in CLIP-based zero-shot image captioning and propose Cross-modal Language Models (CLMs) to address it from unsupervised data in an auto-regressive fashion. By CLMs, the generative model learns cross-modal knowledge through the representation of CLIP.
\item We propose Anchor Augment to further improve the attention of the generative model to the fine-grained information in representation of CLIP. We extract anchors from the original data and use them in CLMs training to guide the generative model.
\item Our approach achieves superior results than existing zero-shot image captioning approaches on both MS COCO and Flickr 30K. In addition, it has a significant advantage in generative speed. 
\end{itemize}

\begin{figure*}
\centering
\includegraphics[width=0.97 \textwidth]{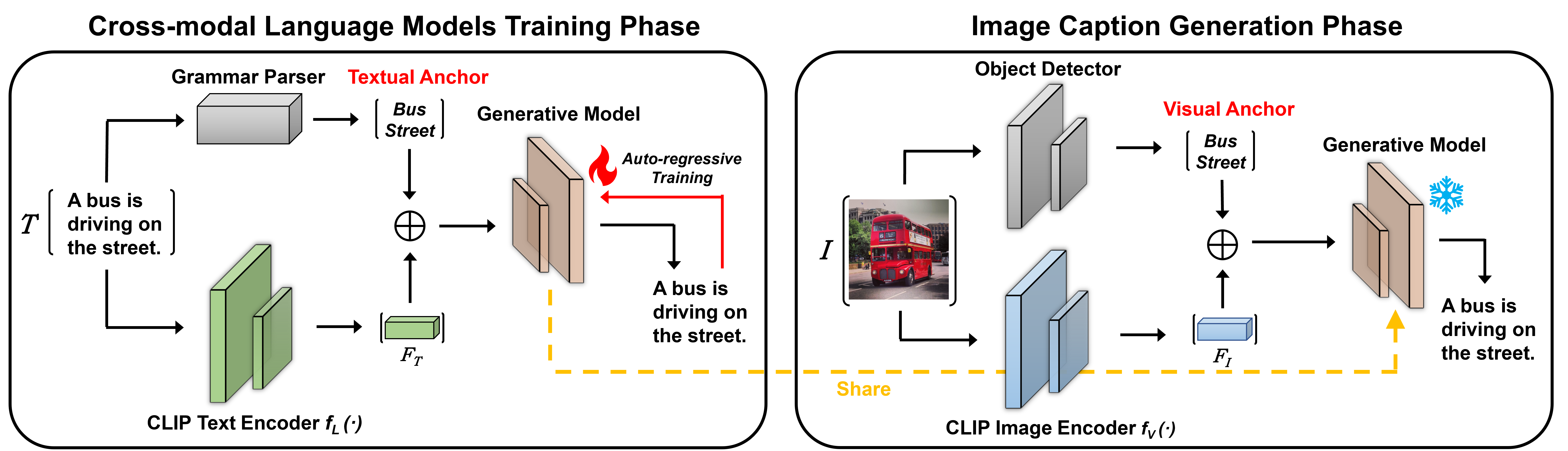}
\caption{\label{fig:approach} The overview of our approach. In the language model training phase, we use Cross-modal Language Models (CLMs) with anchors to train a generative model that can generate the training text based on textual representation extracted by CLIP text encoder and anchors extracted by a grammar parser. The process is self-supervised and without supervised data. In the caption generation phase, we use the trained generative model to generate the caption depending on the visual representation extracted by CLIP image encoder and anchors extracted by a object detector.}
\end{figure*}

\section{Background and Related Work}

\subsection{CLIP}
By contrastive learning on a dataset of 400 million (image, text) pairs, CLIP \cite{radford2021learning} can encode data from visual and textual modalities independently into a common vision-language semantic space. This means that similar image and text are aligned in this space, thus bridging the gap between visual and textual modalities and enabling the learning of transferable cross-modal representation. With the strong generalization obtained in the pre-training phase, CLIP can be used for zero-shot tasks such as classification, retrieval, etc \cite{radford2021learning}. The zero-shot performance is claimed to be close to or even better than fine-tuned models \cite{radford2021learning}. Many works have applied the zero-shot ability of CLIP to specific application scenarios such as image segmentation \cite{xu2021simple}, image generation \cite{ramesh2022hierarchical}, and object detection \cite{zhong2022regionCLIP}. 

\subsection{Zero-Shot Image Captioning}
Early implementations of this task relied on visual features extracted by convolutional neural network\cite{vinyals2015show, xu2015show}. \cite{anderson2018bottom} pioneered the use of an object detector and a bottom-up approach to extract visual features, which lead to a better attention mechanism for the model. With the rise of vision-language pre-training models, researchers have commonly adopted the approach of learning cross-modal universal knowledge by using large-scale image-text pairs in the pre-training phase and fine-tuning on the image captioning datasets\cite{tan2019lxmert, chen2019uniter, li2020oscar, li2021align, dai2022enabling, mokady2021CLIPcap}.

As the development of supervised image captioning is greatly limited due to the high data collection costs, unsupervised approaches attract a lot of attention. 
The powerful zero-shot ability of CLIP has encouraged research on the zero-shot image captioning. 
\cite{tewel2022zerocap} proposes to employ the native GPT-2 \cite{radford2019language} to provide candidate words for the next position based on the context of generated text. The generated captions and each candidate word are then combined into a new set of candidate captions. Finally, the CLIP similarity between these candidate captions and the given image is calculated separately, and the candidate caption with the highest similarity is selected as the newly generated caption at the current position. \cite{su2022language} improves upon this by introducing target domain contextual language prior. They first fine-tune the GPT-2 on the unsupervised caption corpus of the target domain. Then, they combined fine-tuned GPT-2 and CLIP to compute similarity scores for different word candidates for a given image, and selected the word with the highest similarity score as the content of the caption. 

The above works use the GPT-2 to generate candidate words based not on the image but only on the uni-modal context of generated text. This results in heavy influence caused by the \emph{contextual language prior}, which may generate captions that do not correspond to the images. We discuss that, for a cross-modal task such as image captioning, the generative model needs to adequately consider both the image and the context together. This motivates us to achieve a cross-modal understanding of the generative model.

\section{Method}
In this section, we ﬁrst describe the zero-shot image captioning and introduce the vision-language pre-training dual-encoder model CLIP and language model GPT-2 employed by our approach, which are used to extract cross-modal representations and textualize cross-modal representations, respectively. We propose a cross-model language modeling framework, which does not need to use any supervised data in training, and generates captions based on the cross-modal representations extracted by CLIP (Section.\ref{regression}). Based on the framework, we propose Anchor Augment (Section.\ref{anchor}) to improve the attention to the fine-grained information. The overview of our approach is shown in Figure \ref{fig:approach}.


\subsection{Preliminaries}


\textbf{Notations.} 
We first describe supervised data and unsupervised data in the image captioning problem. 
The supervised dataset $\mathcal{D}_{s}$ = \{$(x_1, y_1)$, $\dots$, $(x_n, y_n)$\} consisting of $n$ pairs with images $x_i$ and reference captions $y_i$ = \{$c^1_i$, $\dots$, $c^{|y_i|}_i$\}, where $y_i$ is a set of the captions that describe the $x_i$ from different perspectives, and $c^j_i$ denotes the $j_{th}$ caption of $y_i$. The unsupervised data includes unlabeled image dataset $\mathcal{D}^{I}_{u}$ = \{$x_1$, $\dots$, $x_i$\} and text datasets $\mathcal{D}^{T}_{u}$ = \{$y_1$, $\dots$, $y_j$\}. Traditional image captioning approaches use supervised dataset $\mathcal{D}_{s}$ for training, while zero-shot image captioning only assumes the availability of unlabeled dataset $\mathcal{D}^I_u$ and $\mathcal{D}^T_u$.

\textbf{Base models.}
\textbf{CLIP} is a VLP model with dual-encoder architecture. It consists of two independent encoders for visual and textual modalities. Similarities between vision and language representations on large-scale image-text pairs are used to pre-train CLIP, bridging the gap between vision-language semantics in the representation space of CLIP. 
The similarity is calculated as
\begin{equation}
\begin{aligned}
&F_I = f_{\textup{V}}(I)\\
&F_T = f_{\textup{L}}(T)\\
\textup{Similarity}(I,T) = &\cos<F_I, F_T> = \frac{F_I}{|F_I|}\cdot\frac{F_T}{|F_T|}  
\end{aligned}
\end{equation}
where $f_V$ and $f_L$ is the image encoder and text encoder of CLIP respectively.

\textbf{GPT-2} is a transformer-based language model trained on a large-scale text corpus by an auto-regressive pre-training task. It learns the relationship of sentence context by the next-token prediction. The pre-trained GPT-2 can be fine-tuned on downstream datasets and thus used for continuous specific text generation, such as literature abstract, dialogue, story, etc. The pre-training loss of GPT-2 is the Maximum Likelihood Estimation (MLE) and calculated as
\begin{equation}
\mathcal{L}_{\textup{MLE}} = -\frac{1}{|T|}\sum^{|T|}_{i=1} \log D_\theta(T_i|T_1T_2\ldots T_{i-1})
\end{equation}
where $\theta$ denotes the parameter that needs to be optimized for model $D$. 


\subsection{Cross-modal Language Models by Auto-regressive Training \label{regression}}

The visual and textual representations of CLIP are aligned through the contrastive learning on large-scale image-text pairs.
That is, the textual representation on CLIP can be seen as the visual representation of the matched image on CLIP, even though the CLIP representations of visual modality and textual modality are separately encoded by two uni-modal encoders.
Inspired by this, we propose the Cross-modal Language Models (CLMs) to learn to textualize understanding for visual representations without any image training data. The overview of CLMs training is shown on the left of Figure \ref{fig:approach}. 
  

GPT-2 architecture is employed as the generative model $D_{\theta}$ to generate the textual description for CLIP representations. Based on text encoder $f_{\textup{L}}(\cdot)$ of CLIP, we first obtain the representation $F_T$ of the training text $T$. Then, we put the representation into the position of the first token of GPT-2 and generate the original text by the constraint of auto-regressive loss. 
Note that the token dimension of the GPT-2 architecture we are using is the same as the CLIP representation dimension, which allows the representation of CLIP to be fed into GPT-2 like a normal token. For settings with inconsistent dimensions, an additional adapter module is required to unify the dimensions.
The training loss is formulated as follows
\begin{equation}
\mathcal{L}_{\textup{MLE}} = -\frac{1}{|T|}\sum^{|T|}_{i=1} \log D_\theta(T_i| \pmb{F_T} T_1T_2\ldots T_{i-1})
\end{equation}

The input format is formulated as follows

\begin{equation} \label{text_input}
[cls][F_T][sep][T_1] \ldots [T_n][cls]
\end{equation}
where [$cls$] denotes a special token used at the beginning and the end of the input and [$sep$] denotes a special token for splitting different parts of the input. For example, if the training text is ``A man with a red helmet is riding a motorbike on a dirt road'', the input will be [$cls$][$F_T$][$sep$][$A$][$man$][$with$] \ldots [$a$][$dirt$][$road$][$cls$].

\subsection{Improving the Attention with Anchor Augment\label{anchor}}

To improve the attention of the generative model to the key information in the cross-modal representation of CLIP, we propose Anchor Augment. We want to explicitly supply the generative model with the fine-grained information in the input data so that the generative model can focus more attention on the information. Therefore, we extract the key information from the input text or image as anchors and apply them to the CLMs training. To make the anchors accurate, we directly use a grammar parser to obtain nouns in the training text as anchors as shown in the left of Figure \ref{fig:approach}. After applying the anchors, the training loss is shown by the following equation

\begin{equation}
\mathcal{L}_{\textup{MLE}} = -\frac{1}{|T|}\sum^{|T|}_{i=1} \log D_\theta(T_i| \pmb{F_TA_1 \ldots A_n} T_1\ldots T_{i-1})
\end{equation}
where $A_n$ denotes the $n_{th}$ anchor in the sentence. 

The input format is formulated as follows

\begin{equation}
[cls][F_T][sep][A_1] \ldots [A_n][sep][T_1] \ldots [T_n][cls]
\end{equation}
For example, if the training text is ``A man with a red helmet is riding a motorbike on a dirt road'' whose anchor is ``man'', ``helmat'', ``motorbike'' and ``road'', the input will be [$cls$][$F_T$][$sep$][$man$][$helmat$][$motorbike$][$road$][$sep$][$A$]\\{ }
[$man$][$with$] \ldots [$a$][$dirt$][$road$][$cls$].

In the generation phase, we use an object detector to extract the labels of the image's ROI (Region of Interest). Since the anchors extracted by the object detector are noisy, the model needs to be robust to anchors. To achieve this, we use anchor random dropout. We drop out all anchors with $q$ probability randomly for each training text. In the caption generation phase, we do not use the anchor random dropout.

\subsection{Zero-shot Image Captioning\label{generation}}

The overview of the image caption generation phase is shown on the right of Figure \ref{fig:approach}. Similar to the training phase, the input images are fed into two modules: image encoder $f_{\textup{V}}(\cdot)$ of CLIP and object detector. The image encoder is used to obtain the CLIP representation of the image and the object detector is used to obtain the anchors from the image. For anchor extraction, we set a confidence threshold $p$ for the object detector. Only objects with a confidence greater than $p$ are used for generation. A larger value of $p$ may lose information in the image, while a smaller value of $p$ may introduce noise. After confidence threshold filtering, we extracted the labels of these ROIs as visual anchors. Since the inputs of the training and generation phases are identical in format, we directly share the generative model from CLMs training without any other training. The input format is formulated as follows

\begin{equation}
\label{image_input}
[cls][F_I][sep][A_1] \ldots [A_n][sep]
\end{equation}
We leave the position of the self-supervised label $[T_1] \ldots [T_n][cls]$ empty. Finally, the model generates the image caption in these positions based on the visual representation of CLIP and anchors.

\begin{table*}[t]
	\centering  
	\renewcommand{\arraystretch}{1.2}
	\setlength{\tabcolsep}{8pt}
	\scalebox{0.85}{
	\begin{tabular}{c c c c c c c c c c c c c}
		\hline
		\multirow{2}{*}{\textbf{Approach}}&\multicolumn{6}{c}{MS-COCO}&\multicolumn{6}{c}{Flickr30k}\\
		\cmidrule(lr){2-13}
		&B@1&B@4&M&R-L&CIDEr&SPICE&B@1&B@4&M&R-L&CIDEr&SPICE\\
		\hline
		&\multicolumn{12}{c}{\textit{Weakly Supervised Approach}}\\
		\hline
		UIC \cite{feng2019unsupervised}&41.0&5.6&12.4&28.7&28.6&8.1&-&-&-&-&-&-\\
		IC-SME \cite{laina2019towards}&-&6.5&12.9&35.1&22.7&-&-&7.9&13.0&32.8&9.9&-\\
		S2S-SS \cite{honda2021removing}&49.5&6.3&14.0&34.5&31.9&8.6&-&-&-&-&-&-\\
		S2S-GCC \cite{honda2021removing}&50.4&7.6&13.5&37.3&31.8&8.4&-&-&-&-&-&-\\
		\hline
		&\multicolumn{12}{c}{\textit{Unsupervised Approach}}\\
		\hline
		CLIPRe \cite{su2022language}&39.5&4.9&11.4&29.0&13.6&5.3&38.5&5.2&11.6&27.6&10.0&5.7\\
		ZeroCap \cite{tewel2022zerocap}&49.8&7.0&15.4&31.8&34.5&9.2&44.7&5.4&11.8&27.3&16.8&6.2\\
		SMs \cite{zeng2022socratic}&-&6.9&15.0&34.1&44.5&10.1&-&-&-&-&-&-\\
		MAGIC \cite{su2022language}&56.8&12.9&17.4&39.9&49.3&\textbf{11.3}&44.5&6.4&13.1&31.6&20.4&7.1\\
		\hline
		Ours&\textbf{59.3}&\textbf{15.0}&\textbf{18.7}&\textbf{41.8}&\textbf{55.7}&10.9&\textbf{58.3}&\textbf{16.8}&\textbf{16.2}&\textbf{39.6}&\textbf{22.5}&\textbf{9.8}\\
		\hline
	\end{tabular}
	}
    \caption{Image captioning performances of different approaches on MS-COCO and Flickr30k, where the B@1, B@4, M, and R-L represent BLEU@1, BLEU@4, METEOR, and Rouge-L respectively.}
    	\vspace{-1.5mm}
	\label{tb:main_result}
\end{table*}

\section{Experiment}

In this section, we conduct quantitative and qualitative experiments to evaluate our approach. In the subsection of quantitative experiments, we first compare the performance of our approach with the baselines on various evaluation metrics of image captioning. Then, we conduct ablation experiments to prove the significance of our approach design. Finally, we analyze the effect of the cross-modal transfer by quantitatively observing the performance of the generative model on text generation and image caption generation during the training phase. In the subsection of qualitative experiments, we analyze some example captions generated by the baselines and our approach. 

\textbf{Evaluation Benchmarks.}
We conduct experiments on two widely used benchmarks: MS-COCO \cite{lin2004automatic} and Flickr30k \cite{plummer2015flickr30k}. For both datasets, we set up the training, validation, and test splits according to the splits of Karpathy et al \cite{karpathy2015deep}.

\textbf{Implementation Details.}
We use the text in the training split as corpus for CLMs training. We optimize the generative model with the Adam optimizer \cite{kingma2014adam} and a learning rate of 5e-7. Notably, this procedure is computationally negligible, i.e., less than 3 hours with 1 NVIDIA 3080 GPU. We decide the epoch number of training based on the performance of the model on the validation set. For object detector, we choose the mainstream model Faster-RCNN \cite{ren2015faster}. For the confidence threshold $p$ of the object detector, we chose three values of 0.5, 0.7, and 0.9. For the probability $q$ of anchor random dropout, we chose 0, 0.25, 0.5, 0.75, and 1. For the approach of search, we choose beam search, where the branches of the beam are chosen as 5 which is as same as other approaches.

\textbf{Baselines.}
First we choose some weakly supervised approaches UIC \cite{feng2019unsupervised}, IC-SME \cite{laina2019towards}, S2S-SS and S2S-GCC \cite{honda2021removing}. The starting point of these approaches is to address the problem of data limitation, which is similar to the unsupervised approaches. However, the weakly supervised approaches still require an amount of supervised data to generate unsupervised data. Then we compare with a CLIP-based approach, called CLIPRe. Given an image, it retrieves the most related caption from a caption corpus based on the image-text similarity as measured by CLIP. We also compare three unsupervised approaches from the last year, ZeroCap \cite{tewel2022zerocap}, MAGIC \cite{su2022language}, and SMs \cite{zeng2022socratic}. 

\textbf{Evaluation Metrics.}
Following the common practice in the literature, we perform evaluation using BLEU-1 (B@1), BLEU-4 (B@4) \cite{papineni2002bleu}, METEOR (M) \cite{denkowski2014meteor}, ROUGE-L (R-L) \cite{lin2004automatic}, CIDEr \cite{vedantam2015cider}.

\begin{table*}[h]
    \small
	\centering  
	\renewcommand{\arraystretch}{1.2}
	\setlength{\tabcolsep}{8pt}
	\scalebox{0.93}{
	\begin{tabular}{ccccccccccccc}
	\hline
		\multirow{2}{*}{\textbf{Model}}&\multicolumn{6}{c}{MS-COCO $\Longrightarrow$ Flickr30k}&\multicolumn{6}{c}{Flickr30k $\Longrightarrow$ MS-COCO}\\
		\cmidrule(lr){2-7}
		\cmidrule(lr){8-13}
		&B@1&B@4&M&R-L&CIDEr&SPICE&B@1&B@4&M&R-L&CIDEr&SPICE\\
		\hline
		CLIPRe&38.7&4.4&9.6&27.2&5.9&4.2&31.1&3.0&9.9&22.8&8.5&3.9\\
		MAGIC&\textbf46.4&6.2&12.2&31.3&\textbf{17.5}&\textbf{5.9}&41.4&5.2&12.5&30.7&18.3&5.7\\
		\hline
		Ours&\textbf{49.2}&\textbf{10.1}&\textbf{12.5}&\textbf{33.8}&12.7&5.7&\textbf{47.6}&\textbf{7.7}&\textbf{14.9}&\textbf{35.9}&\textbf{38.5}&\textbf{8.2}\\
		\hline
	\end{tabular}}
    \caption{Cross-Domain Evaluation. X $\Longrightarrow$ Y  means source domain $\Longrightarrow$ target domain.}
    	\vspace{-1.5mm}
	\label{tb:cross_domain_result}
\end{table*}

\begin{table}[t]
	\centering  
	\renewcommand{\arraystretch}{1.2}
	\setlength{\tabcolsep}{9pt}
	\scalebox{0.93}{
	\begin{tabular}{c | c}
		\hline
		Model&Speed (second)\\
		\hline
		ZeroCap&76.8\\
		MAGIC&2.89\\
		\hline
		Ours&1.46 (OD Phase) + 0.27 (ICG Phase)\\
		\hline
	\end{tabular}
	}
    \caption{The caption generation speed of different approaches on a sample. We count the speed of our approach for the object detection (OD) phase and image caption generation (ICG) phase separately.}
    \vspace{-1.5mm}
	\label{tb:speed_result}
\end{table}

\begin{table*}[t]
	\centering  
	\renewcommand{\arraystretch}{1.2}
	\setlength{\tabcolsep}{10pt}
	\scalebox{0.93}{
	\begin{tabular}{c c c c c c c c c c c}
		\hline
		&\multicolumn{5}{c}{MS-COCO}&\multicolumn{5}{c}{Flickr30k}\\
		\hline
		$q$&0&0.25&0.5&0.75&1&0&0.25&0.5&0.75&1\\
		\hline
		$p$ = 0.5&31.7&32.9&32.5&32.0&\multirow{4}{*}{24.6}&19.0&26.5&27.1&27.0&\multirow{4}{*}{26.4}\\
		$p$ = 0.7&32.3&33.7&33.6&33.0&&18.6&25.2&27.2&26.8&\\
		$p$ = 0.9&31.8&33.4&33.5&32.9&&18.7&25.2&27.2&26.8&\\
		$p$ = 1&11.6&20.9&22.3&22.0&&18.0&25.8&26.7&26.6&\\
		\hline
	\end{tabular}
	}
    \caption{We have chosen different values of $p$ and $q$ for the ablation analysis of our approach. The values in the table represent the average of all evaluated metrics. $p$ represents the confidence threshold of the object detector. The objects with confidence greater than this threshold are used as anchors. Where $p$ = 1 means that no anchor is provided in the generation phase. $q$ represents the probability of anchor random dropout in the training phase. $q$ = 1 means that no anchor is added in the training phase. In order to align the settings of the two phases, the generation phase in this case also does not provide anchors, so the value is independent of $p$.}
    	\vspace{-1.5mm}
	\label{tb:ablation}
\end{table*}

\subsection{Quantitative Experiments}

\subsubsection{Image Captioning}
Table \ref{tb:main_result} shows the results on zero-shot image captioning. First, we observe that the unsupervised approaches based on CLIP have better performances compared to weak supervision approaches. This is because the weak supervision approaches need to expand the amount of data according to a small amount of supervision data. Therefore, the weak supervision approach does not really get rid of the data limitations. Second, we observe that the performances of generation-based approaches (ZeroCap, MAGIC, and ours) are far better than the retrieval-based approach (CLIPRe). The retrieval-based approach is limited by the capacity of the text corpus, so it can hardly generate the caption for unseen samples. There is also a problem with the supervision approaches, but the supervision approaches can make the model fit the data so that the model has a certain generalization of the unseen sample. Finally, our approach achieved excellent performance of all approaches, with 11 of its 12 metrics being the best. Not only that, we have a significant improvement on some metrics (such as the B@4 and SPICE on Flickr30K). This is due to the fact that our approach makes the generative model truly attend to the content in the image through cross-modal learning, unlike other approaches that are based on contextual language prior.

\textbf{Generalization.} We also focus on the generalization ability. The result of the cross-domain evaluation is shown in Table \ref{tb:cross_domain_result}. We apply the generative model trained on the training text corpus of the source domain to perform inference on the test set of the target domain. The experimental results show that our approach has a strong cross-domain ability.

\textbf{Efficiency.} We also pay attention to generative efficiency. The experimental results of generative efficiency are shown in Table \ref{tb:speed_result}. We can observe that our approach is much faster than the ZeroCap. That is because ZeroCap involves computationally inefficient operations like gradient updates \cite{dathathri2019plug, tewel2022zerocap}. This efficiency seriously limits its actual application of it. We observe that our approach is more efficient than baselines and the time cost in the image caption generation phase is almost negligible. The efficiency of our approach is mainly limited by the object detector. With the development of faster object detectors, our approach holds promise for use in real-time scenarios.

\subsubsection{Ablation Experiments}

Table \ref{tb:ablation} shows the results of the ablation experiments. We probe the influence of anchors on the model by setting the confidence threshold $p$ of the object detector and the probability $q$ of anchor random dropout. For the convenience of observation, we calculate the mean values of the 6 image captioning evaluation metrics.

\textbf{Confidence Threshold of Object Detector.} First, we probe the influence of $p$ by comparing the value of the same column. From the results, we can see that the performance difference is not significant at $p$ equal to 0.5, 0.7, and 0.9. This verifies that the model is robust to the anchors by anchor random dropout. Even if we provide noisy anchors by a low $p$ for the model in the generation phase, the model is still able to select the ones relevant to the CLIP representation from these anchors. This is because anchor random dropout makes it mandatory for the model to learn the connection between CLIP representation and anchors. However, when we set $p$ to 1, i.e., we do not provide any anchor to the model in the generation phase, the performance of the model will be significantly declined. This shows that using anchors only in the training phase does not effectively improve the performance of the model.

\textbf{Probability of Anchor Random Dropout.} Second, We probe the influence of $q$ by comparing the value of the same line, where $q$ equals 0 means no anchor dropout is used. It is worth noting that $q$ equals 1 means that there is no anchor to participate in training. In this case, we no longer provide anchors for the model in the generation stage. We use a value to represent the performance because in this case, the performance is independent of $p$. We observe that both without the use of anchor random dropout and without anchors involved in the training phase, the performance decreased. This illustrates both the need for the existence of anchor augmentation and the need for caution in the use of anchors. When there is no anchor, the model can only extract information from the CLIP representation that has a low dimension compared to a sentence, which increases the difficulty of generation. And when anchors are fully used, the model creates a dependency on the anchors and thus ignores the information in the CLIP representation. Therefore, an appropriate dropout probability is particularly important. From the results, a dropout probability of 0.5 is a desirable setting. With this setting, the model is able to learn the information in CLIP representation and anchors in a balanced way.

\begin{figure}
\centering
\includegraphics[width=0.46 \textwidth]{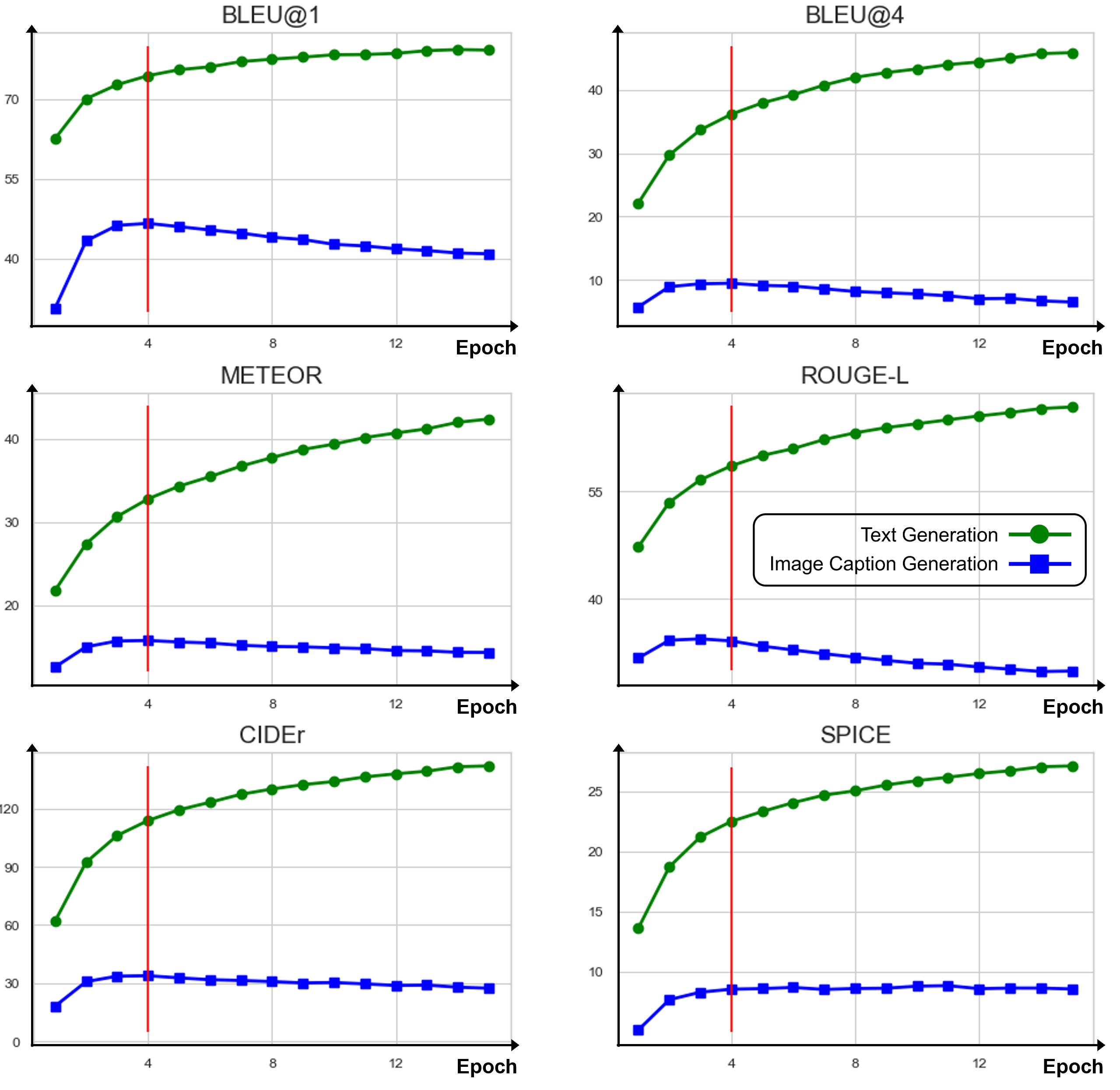}
\caption{\label{fig:train} The performance for text generation and image caption generation varies with the number of training epochs.}
\end{figure}

\begin{figure*}
\centering
\includegraphics[width=0.97 \textwidth]{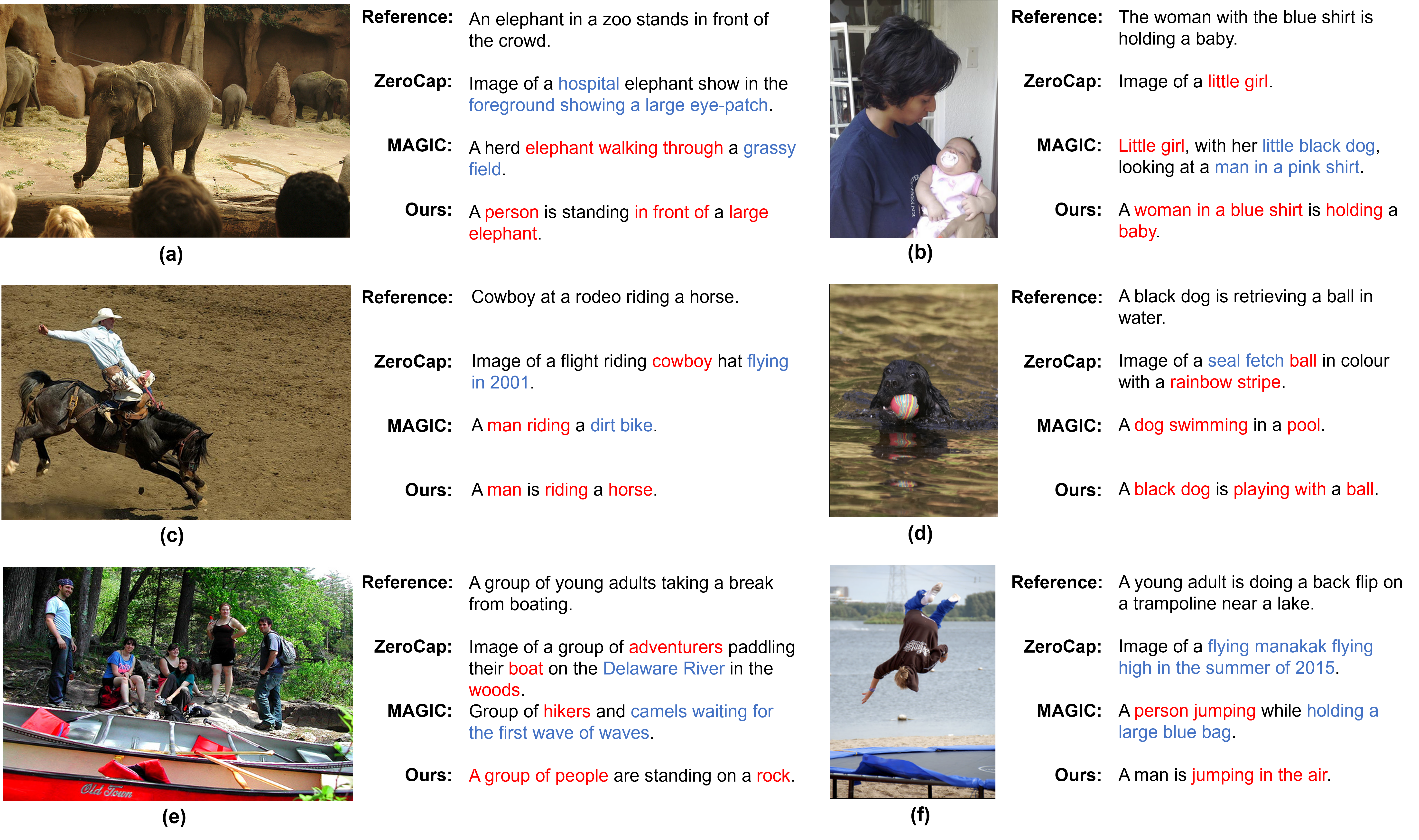}
\caption{\label{fig:intro} We show more captions generated by three zero-shot image captioning approaches: ZeroCap, MAGIC, and ours. The red-colored words refer to relevant information, while the blue-colored words refer to irrelevant information that cannot be inferred from the image.}
\end{figure*}

\begin{figure}
\centering
\includegraphics[width=0.48 \textwidth]{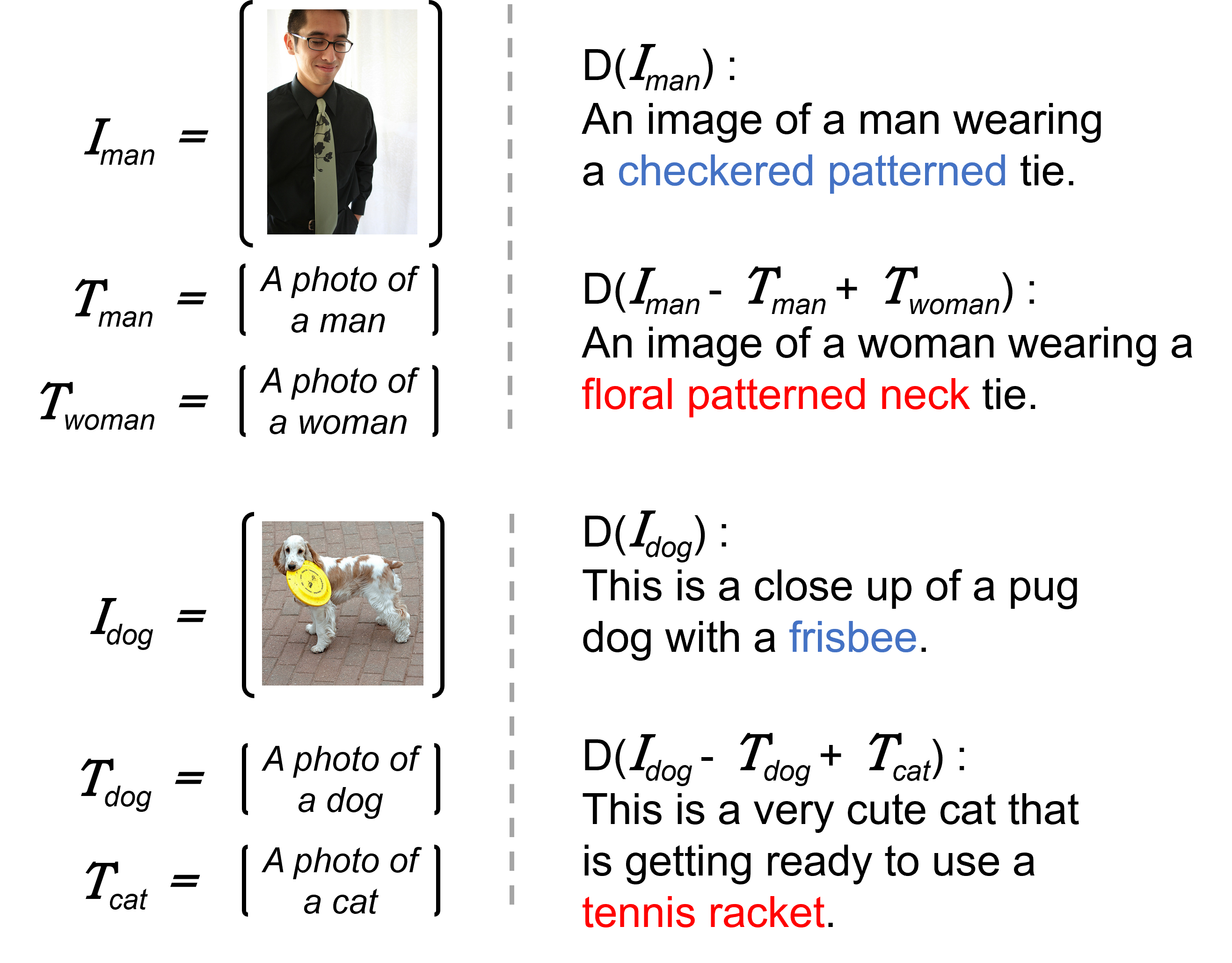}
\caption{\label{fig:discuss} Illustration for explainable bias study by cross-modal counterfactual operation. The blue and red font represent the difference between the captions generated before and after counterfactual operation, respectively.}
\end{figure}

\subsubsection{Analysis of Cross-modal Transfer}


Without the supervised data, we achieve the cross-modal transfer of knowledge through the vision-language space of CLIP. However, the vision-language space of CLIP has been shown to contain biases \cite{berg2022prompt, dehouche2021implicit, agarwal2021evaluating}. We hope the cross-modal transfer to gravitate towards non-biased knowledge as possible. Therefore, we wanted to understand the influence of CLMs training on the cross-modal transfer of knowledge. We apply the generative model for text generation and image caption generation, respectively, where the input format for image caption generation is consistent with Equation \ref{image_input} and for text generation is the replacement of $F_I$ with $F_T$ in Equation \ref{image_input}. We calculate the generation quality for both cases separately based on the reference caption and plot line graphs of the different metrics with the epochs of training. The results are shown in Figure \ref{fig:train}. We observe that at the early stage of training (before the red line), CLMs training greatly promotes the cross-modal transfer of knowledge. However, as the training progresses (after the red line), CLMs inhibit cross-modal transfer. This means we need to stop CLMs training at an appropriate point.

\subsection{Qualitative Experiments}

Figure \ref{fig:intro} shows visual comparisons between our approach and the two zero-shot baselines along with the reference caption. The results demonstrate that our approach can generate fluent captions. 

We can clearly see that ZeroCap is almost failing to generate quality captions. The generated captions contain a lot of unsupported content. Although the authors argue that ZeroCap's generation style makes sense for unsupervised image captioning because it can greatly enrich the diversity of captions, we believe that such unsupported content is somewhat risky. For example, by inputting an image, ZeroCap runs the risk of generating some private content, such as the location where it was taken. 

The reason why MAGIC and our approach are much better in this respect is that we use clean text for training, which allows effective control over the generation style of the generative model. However, MAGIC's lack of cross-modal learning makes its language model only a uni-modal model. This means that the caption generated by MAGIC contains severe context language prior. For example in Figure \ref{fig:intro} a, there is a context language prior between the elephant and the grass in the language training corpus, which is also reflected in the generated results. When the model generated the ``elephant'', the subsequent generation did not consider the fact that the elephant is standing on the ground in the image but generated the ``grassy'' based on the experience of the language model. In all of the examples of Figure \ref{fig:intro}, our approach achieves accurate generation. In addition, our approach is more accurate in identifying objects. For example, in Figure \ref{fig:intro} b, our approach accurately identifies the ``baby'' that is not identified by other approaches. This is made possible by the use of anchors.

\section{Discussion}

Currently, CLIP has been widely used as the base model for various tasks. All along, researchers have tried to figure out what CLIP learned from the web data. \cite{bolukbasi2016man} observed that a subspace of concepts can be obtained by performing linear operations on the word vectors in the embedding space. We first use this property to perform a linear operation on the textual representation of CLIP to obtain a subspace that represents a concept change to another concept. Then, by applying this subspace to the visual representation of an image, a counterfactual representation is obtained. We use the example of detecting the stereotypes in CLIP as shown in Figure \ref{fig:discuss}. First, we obtain the subspace of \textit{man} to \textit{woman} and \textit{dog} to \textit{cat} by linear operations. Then, we generate captions for the original representations and counterfactual representations, respectively. We observe that there are attributes that should be not related to the concept of subspace change. For example, when we change the gender concept in the image from \textit{man} to \textit{woman}, the \textit{checkered} change to \textit{floral}, even though the pattern of the tie should be not related to the gender. This suggests spurious stereotypes in CLIP.
Accordingly, our approach can be used to assess/interpret dependencies and connections between concepts in CLIP-style models, and can be generalized as a tool to provide new insights to explore the modeling of information in opaque VLP models.

\section{Conclusion}

In this paper, we propose Anchor-augmented Vision-Language Space Alignment for zero-shot image captioning. 
To avoid the unintentional introduction of contextual language priors caused by uni-modal language models in previous approaches, we propose Cross-modal Language Models training task.
In addition, to improve the attention of the generative model to the fine-grained information in the cross-modal representation of CLIP, we propose Anchor Augment. The training process of our approach is efficient and does not require annotated data. The experiment results demonstrate that our approach can achieve SOTA in generation quality compared with other baselines. Also because of the non-query-based architecture, our approach achieves a high generation efficiency. 

\clearpage
{\small
\bibliographystyle{ieee_fullname}
\bibliography{egbib}
}

\end{document}